\documentclass{midl} % Include author names
\usepackage{mwe} % to get dummy images
\usepackage{hhline} % to get dummy images

\jmlrproceedings{MIDL 2020}{Medical Imaging with Deep Learning 2020}
\jmlrvolume{}
\jmlryear{}
\jmlrworkshop{MIDL 2020 -- Short Paper}
\editors{}

\title[Unsupervised multimodal registration using Earth Mover's discrepancies]{Unsupervised learning of multimodal image registration using domain adaptation with projected Earth Mover's discrepancies}

\midlauthor{\Name{Mattias P. Heinrich} \Email{heinrich@imi.uni-luebeck.de}\AND\Name{Lasse Hansen} \Email{hansen@imi.uni-luebeck.de}\\
\addr  Institute of Medical Informatics, Universit\"at zu L\"ubeck, L\"ubeck, Germany \\
}

\begin{document}

\maketitle

\begin{abstract}
Multimodal image registration is a very challenging problem for deep learning approaches. Most current work focuses on either supervised learning that requires labelled training scans and may yield models that bias towards annotated structures or unsupervised approaches that are based on hand-crafted similarity metrics and may therefore not outperform their classical non-trained counterparts. We believe that unsupervised domain adaptation can be beneficial in overcoming the current limitations for multimodal registration, where good metrics are hard to define.
Domain adaptation has so far been mainly limited to classification problems. We propose the first use of unsupervised domain adaptation for discrete multimodal registration. Based on a source domain for which quantised displacement labels are available as supervision, we transfer the output distribution of the network to better resemble the target domain (other modality) using classifier discrepancies. To improve upon the sliced Wasserstein metric for 2D histograms, we present a novel approximation that projects predictions into 1D and computes the L1 distance of their cumulative sums. Our proof-of-concept demonstrates the applicability of domain transfer from mono- to multimodal (multi-contrast) 2D registration of canine MRI scans and improves the registration accuracy from 33\% (using sliced Wasserstein) to 44\%. 
\end{abstract}
% and in general regression of continuous variables is a harder optimisation task in deep learning (therefore localisation is often posed as discrete heatmap estimation)
\begin{keywords}
Multi-modal registration, domain adaptation, discrete displacements
\end{keywords}

\section{Introduction}
Gathering labelled training data for learning-based multimodal registration is very time-consuming and expensive. To train supervise methods either a large number of corresponding landmarks (cf. \cite{xiao2019evaluation}) or detailed anatomical multi-label segmentations are required (cf. \cite{hu2018weakly}), which often cause bias or under-coverage. To circumvent the need for corresponding labels in multimodal / multi-domain images, unsupervised domain adaptation based on classifier discrepancies has been popularised in computer vision for classification and segmentation tasks e.g. in \cite{lee2019sliced}. Variants of discrepancy measures include e.g. the Earth Mover's distance (EMD) \cite{werman1985distance} for 1D cases and specialised solutions for 2D histograms in \cite{ling2006diffusion}, but they are in general computationally expensive, approximative or based on sensitive hyperparameters.
%In addition currently available registration networks are commonly pre-trained using either simulated deformations (Eppenhof) or based on labelled atlases (Dalca), which are available only within the same modality. 

\textbf{Contribution:} We are the first to propose domain adaptation for medical registration and adapt the task to a discrete displacement labelling. Using the maximum classifier discrepancy approach \cite{saito2018maximum} together with a novel 2D histogram Earth Movers distance, we substantially improve over the  sliced Wasserstein metric \cite{lee2019sliced}.   

\textbf{Related Work:}
Recent methods for supervised learning of multimodal registration include \cite{simonovsky2016deep}, who use a twin CNN architecture to learn the similarity of patches using aligned multi-modal training data. \cite{hu2018weakly} and \cite{hering2019memory}, both use anatomical segmentations  to train a U-net like registration network, while the latter add a normalised gradient metric. The use of discrete displacements in deep learning based registration was proposed in \cite{heinrich2019closing} to capture large deformations.
%\cite{blendowski2019learning} learn a CNN feature extractor through guidance of an iterative registration framework that is supervised by label images. 
Unpaired unsupervised learning for multi-modal medical images has so far been restricted to modality synthesis using e.g. Cycle-GANs in \cite{wolterink2017deep}. Very recent methods have shown promise for unsupervised domain adaptation and knowledge distillation for medical image classification and multimodal segmentation \cite{dou2020unpaired}.

\section{Methods and Material:}
Unsupervised domain adaption has so far mainly shown success for classification tasks. We hence adapt the task of multimodal image registration to a discrete labelling problem, similar as done in \cite{heinrich2019closing}. Here, we restrict ourselves to 2D patch based registration to demonstrate a proof-of-concept. For training, we extract large patches with a random offset within a grid of 5x5 discrete displacements to pose registration as a 25-class classification problem. We add 3D affine augmentations to avoid trivial overlap within two patches. 

During training we have access to a labelled source domain dataset (in this case MRI T1 patches) with known displacements and an unlabelled target domain dataset (MRI T2 patches). For feature extraction, we use a feed-forward net comprising four blocks of Conv2d, InstanceNorm and PReLU (13k weights) within a twin architecture that shares weights across both patches. This feature network produces a 18x18 map with 16 channels. Subsequently, we concatenate both patches and feed them into a three block classification network (70k weights) that predicts a 25D classification vector (encoding the displacements). 

\begin{figure}[tb]
 % Caption and label go in the first argument and the figure contents
 % go in the second argument
\floatconts
  {fig1}
  {\caption{Our method comprises a shared feature network and two classifiers for maximum discrepancy domain adaption. The results demonstrate the superiority of our new p-EMD metric (44\% vs 33\% accuracy) compared to sliced Wasserstein (SWD). }}
  {\includegraphics[width=1\linewidth]{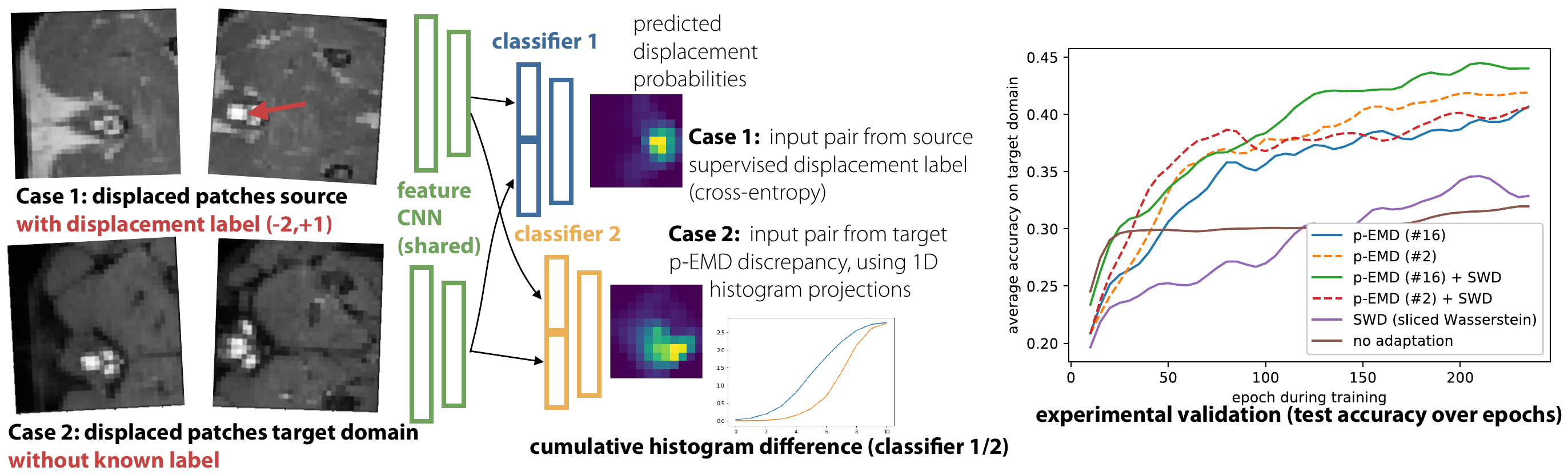}}
\end{figure}

We employ the maximum discrepancy of classifiers approach of \cite{saito2018maximum}, which uses two similar classifiers (with different random initialisation). The training alternates between the following three steps (see also Figure \ref{fig1}): 1) optimise features and classifier on labelled source data, 2) maximise discrepancy measure of both classifiers on target domain while freezing the feature weights and minimising the classification loss on the labelled source data, 3) minimise the discrepancy measure of classifiers while updating only the feature weights.
This process helps to identify target samples outside the classifier's support region in step 2 and brings the target feature distributions closer to the source ones during step 3. Step 3 is repeated twice as proposed in \cite{saito2018maximum}. We make two modifications that greatly improve stability: 1) we only update the first classifier in step 1 to avoid overfitting (too similar decision boundaries) on the source domain before the domain adaptation begins to improve, 2) we use the cross entropy loss for classifiers, but scale the predictions by 0.1 before computing the softmax output for the discrepancy measures to reduce overly confident predictions as motivated by \cite{kuleshov2018accurate}.% on Platt scaling for reliable uncertainty estimation. 
%An obvious choice for the discrepancy measure would be the sliced Wasserstein metric that set new state-of-the-art accuracies for unsupervised domain adaption in classification and segmentation of natural images \cite{lee2019sliced}.

\textbf{Fast projected Earth Mover's distance (p-EMD) for multidimensional histograms:}
A disadvantage of the sliced Wasserstein distance \cite{lee2019sliced} for our application is its invariance to permutations of histogram bins / classes. This may be beneficial when no natural measure of class proximity exists. Yet, in our case the prediction can be converted into a 2D spatial probability map for x- and y-displacements. We therefore propose a new approximate metric for higher-order histograms (p-EMD) that takes these specificities into account. It is faster and easier to differentiate than conventional algorithms. Given that the distributions are close to monomodal Gaussians and based on the fact that exact algorithms for computing EMD for normalised 1D histograms in linear complexity exist \cite{werman1985distance}, we approximate the optimal transport cost by projecting the (softmax) normalised 2D histograms onto a number of rotated lines (we use either 2 or 16 projections with angles between 0 and 90 degrees and use bilinear interpolation). We then employ the L1 distance of their cumulative sums to compute the p-EMD and average the values across projections (see Figure \ref{fig1}), this correlates nearly perfectly with exhaustive EMD computations and is much more stable in our experiments than the 2D diffusion distance of \cite{ling2006diffusion}.
\section{Results and Discussion}
We created a multimodal dataset for patch registration based on 9 3D T1 and T2 MRI scans of canine legs as provided by the 2013 MICCAI SATA challenge \cite{asman2013miccai} with 5120 patch pairs in each modality. T1to T2 MRI is a simpler domain adaptation task, we thus increase the complexity by applying slightly different normalisations to the patches (global mean and variance for T1, and patch-wise for T2). The range of displacements was $\{-38,-19,0,+19,+38\}^2$ pixels (posing a very challenging large motion problem) and each patch comprises a region of 77x77 voxels downsampled to half resolution. The supervised training was restricted to monomodal data (T1$\rightarrow$T1), while the multi-modal tests were performed on T2$\rightarrow$T1, T1$\rightarrow$T2 and T2$\rightarrow$T2. The average accuracy (prediction of 1 of the 25 classes) across 5 runs is shown in Fig.~\ref{fig1} (right), yielding only a modest improvement from 31.9\% (no adaptation) for sliced Wasserstein (SWD) to 33.2\%. Both of our variants p-EMD (2 or 16 projections) reach accuracies over 40\%, adding the losses of p-EMD (\#16) and SWD is best with 44.1\% (see also Table~\ref{summary}). Future work will focus on more elaborate experiments and evaluation, e.g. integrating the patch-wise displacement estimation into global transformation models (e.g. using the instance optimisation proposed in \cite{heinrich2019closing}), extending it to 3D and comparison to classical multimodal metrics.

% Acknowledgments---Will not appear in anonymized version
\midlacknowledgments{This work was in part supported by the German ministry of Education and Research (BMBF) within the project Multi-Task Deep Learning for Large-Scale Multimodal Biomedical Image Analysis (MDLMA) FKZ 031L0202B.}

\begin{table}[htp]
\caption{Overview of label accuracy results for multi-modal MRI registrations}
\begin{center}
\begin{tabular}{c|c}
\hline
method&registration label accuracy\\
\hhline{=|=} 
no registration (guessing)&4.0\%\\
\hline
no adaptation (training on T1)& 31.9\%\\
\hline
sliced Wasserstein (SWD) & 33.2\%\\
\hline
p-EMD (\#16) and SWD (ours)& 44.1\%\\
\hline

\end{tabular}
\end{center}
\label{summary}
\end{table}%
\bibliography{midl-samplebibliography}

\begin{thebibliography}{13}
\providecommand{\natexlab}[1]{#1}
\providecommand{\url}[1]{\texttt{#1}}
\expandafter\ifx\csname urlstyle\endcsname\relax
  \providecommand{\doi}[1]{doi: #1}\else
  \providecommand{\doi}{doi: \begingroup \urlstyle{rm}\Url}\fi

\bibitem[Asman et~al.(2013)Asman, Akhondi-Asl, Wang, Tustison, Avants,
  Warfield, and Landman]{asman2013miccai}
Andrew Asman, Alireza Akhondi-Asl, Hongzhi Wang, Nicholas Tustison, Brian
  Avants, Simon~K Warfield, and Bennett Landman.
\newblock Miccai 2013 segmentation algorithms, theory and applications (sata)
  challenge results summary.
\newblock In \emph{MICCAI Challenge Workshop on Segmentation: Algorithms,
  Theory and Applications (SATA)}, 2013.

\bibitem[Dou et~al.(2020)Dou, Liu, Heng, and Glocker]{dou2020unpaired}
Qi~Dou, Quande Liu, Pheng~Ann Heng, and Ben Glocker.
\newblock Unpaired multi-modal segmentation via knowledge distillation.
\newblock \emph{arXiv preprint arXiv:2001.03111}, 2020.

\bibitem[Heinrich(2019)]{heinrich2019closing}
Mattias~P Heinrich.
\newblock Closing the gap between deep and conventional image registration
  using probabilistic dense displacement networks.
\newblock In \emph{International Conference on Medical Image Computing and
  Computer-Assisted Intervention}, pages 50--58. Springer, 2019.

\bibitem[Hering et~al.(2019)Hering, Kuckertz, Heldmann, and
  Heinrich]{hering2019memory}
Alessa Hering, Sven Kuckertz, Stefan Heldmann, and Mattias~P Heinrich.
\newblock Memory-efficient 2.5 d convolutional transformer networks for
  multi-modal deformable registration with weak label supervision applied to
  whole-heart ct and mri scans.
\newblock \emph{International journal of computer assisted radiology and
  surgery}, 14\penalty0 (11):\penalty0 1901--1912, 2019.

\bibitem[Hu et~al.(2018)Hu, Modat, Gibson, Li, Ghavami, Bonmati, Wang, Bandula,
  Moore, Emberton, et~al.]{hu2018weakly}
Yipeng Hu, Marc Modat, Eli Gibson, Wenqi Li, Nooshin Ghavami, Ester Bonmati,
  Guotai Wang, Steven Bandula, Caroline~M Moore, Mark Emberton, et~al.
\newblock Weakly-supervised convolutional neural networks for multimodal image
  registration.
\newblock \emph{Medical image analysis}, 49:\penalty0 1--13, 2018.

\bibitem[Kuleshov et~al.(2018)Kuleshov, Fenner, and
  Ermon]{kuleshov2018accurate}
Volodymyr Kuleshov, Nathan Fenner, and Stefano Ermon.
\newblock Accurate uncertainties for deep learning using calibrated regression.
\newblock \emph{arXiv preprint arXiv:1807.00263}, 2018.

\bibitem[Lee et~al.(2019)Lee, Batra, Baig, and Ulbricht]{lee2019sliced}
Chen-Yu Lee, Tanmay Batra, Mohammad~Haris Baig, and Daniel Ulbricht.
\newblock Sliced wasserstein discrepancy for unsupervised domain adaptation.
\newblock In \emph{Proceedings of the IEEE Conference on Computer Vision and
  Pattern Recognition}, pages 10285--10295, 2019.

\bibitem[Ling and Okada(2006)]{ling2006diffusion}
Haibin Ling and Kazunori Okada.
\newblock Diffusion distance for histogram comparison.
\newblock In \emph{2006 IEEE Computer Society Conference on Computer Vision and
  Pattern Recognition (CVPR'06)}, volume~1, pages 246--253. IEEE, 2006.

\bibitem[Saito et~al.(2018)Saito, Watanabe, Ushiku, and
  Harada]{saito2018maximum}
Kuniaki Saito, Kohei Watanabe, Yoshitaka Ushiku, and Tatsuya Harada.
\newblock Maximum classifier discrepancy for unsupervised domain adaptation.
\newblock In \emph{Proceedings of the IEEE Conference on Computer Vision and
  Pattern Recognition}, pages 3723--3732, 2018.

\bibitem[Simonovsky et~al.(2016)Simonovsky, Guti{\'e}rrez-Becker, Mateus,
  Navab, and Komodakis]{simonovsky2016deep}
Martin Simonovsky, Benjam{\'\i}n Guti{\'e}rrez-Becker, Diana Mateus, Nassir
  Navab, and Nikos Komodakis.
\newblock A deep metric for multimodal registration.
\newblock In \emph{International conference on medical image computing and
  computer-assisted intervention}, pages 10--18. Springer, 2016.

\bibitem[Werman et~al.(1985)Werman, Peleg, and Rosenfeld]{werman1985distance}
Michael Werman, Shmuel Peleg, and Azriel Rosenfeld.
\newblock A distance metric for multidimensional histograms.
\newblock \emph{Computer Vision, Graphics, and Image Processing}, 32\penalty0
  (3):\penalty0 328--336, 1985.

\bibitem[Wolterink et~al.(2017)Wolterink, Dinkla, Savenije, Seevinck, van~den
  Berg, and I{\v{s}}gum]{wolterink2017deep}
Jelmer~M Wolterink, Anna~M Dinkla, Mark~HF Savenije, Peter~R Seevinck,
  Cornelis~AT van~den Berg, and Ivana I{\v{s}}gum.
\newblock Deep mr to ct synthesis using unpaired data.
\newblock In \emph{International workshop on simulation and synthesis in
  medical imaging}, pages 14--23. Springer, 2017.

\bibitem[Xiao et~al.(2019)Xiao, Rivaz, Chabanas, Fortin, Machado, Ou, Heinrich,
  Schnabel, Zhong, Maier, et~al.]{xiao2019evaluation}
Yiming Xiao, Hassan Rivaz, Matthieu Chabanas, Maryse Fortin, Ines Machado,
  Yangming Ou, Mattias~P Heinrich, Julia~A Schnabel, Xia Zhong, Andreas Maier,
  et~al.
\newblock Evaluation of mri to ultrasound registration methods for brain shift
  correction: The curious2018 challenge.
\newblock \emph{IEEE Transactions on Medical Imaging}, 2019.

\end{thebibliography}

\end{document}